\documentclass[letterpaper, 10 pt, conference]{ieeeconf}  %

\IEEEoverridecommandlockouts                              %

\usepackage[T1]{fontenc}
\usepackage{balance}
\usepackage{amsmath}
\usepackage{amssymb}
\usepackage{graphicx}
\usepackage{xspace}

\usepackage{times}
\usepackage[dvipsnames]{xcolor}
\usepackage{colortbl}
\usepackage{xurl}
\usepackage{booktabs}  %

\makeatletter
\let\NAT@parse\undefined
\makeatother
\usepackage[numbers,sort&compress]{natbib}
\definecolor{cvprblue}{rgb}{0.21,0.49,0.74}
\usepackage[breaklinks,colorlinks,citecolor=cvprblue]{hyperref}

\makeatletter
\DeclareRobustCommand\onedot{\futurelet\@let@token\@onedot}
\def\@onedot{\ifx\@let@token.\else.\null\fi\xspace}

\def\eg{\emph{e.g}\onedot} 
\def\ie{\emph{i.e}\onedot}

\makeatother

\title{\LARGE \bf
\textbf{SafeFlow}: Real-Time Text-Driven Humanoid Whole-Body Control via\\Physics-Guided Rectified Flow and Selective Safety Gating
}

\author{
Hanbyel Cho \quad Sang-Hun Kim \quad Jeonguk Kang \quad Donghan Koo \\
Physical AI Lab, RX, Samsung Electronics \\
Project page: \href{https://hanbyelcho.info/safeflow/}
{\textcolor{cvprblue}{https://hanbyelcho.info/safeflow/}}
}

\begin{document}
\maketitle
\thispagestyle{empty}
\pagestyle{empty}

\begin{abstract}
Recent advances in real-time interactive text-driven motion generation have enabled humanoids to perform diverse behaviors. However, kinematics-only generators often exhibit physical hallucinations, producing motion trajectories that are physically infeasible to track with a downstream motion tracking controller or unsafe for real-world deployment. These failures often arise from the lack of explicit physics-aware objectives for real-robot execution and become more severe under out-of-distribution (OOD) user inputs. Hence, we propose \textbf{SafeFlow}, a text-driven humanoid whole-body control framework that combines physics-guided motion generation with a 3-Stage Safety Gate driven by explicit risk indicators. \textbf{SafeFlow} adopts a two-level architecture. At the high level, we generate motion trajectories using Physics-Guided Rectified Flow Matching in a VAE latent space to improve real-robot executability, and further accelerate sampling via Reflow to reduce the number of function evaluations (NFE) for real-time control. The 3-Stage Safety Gate enables selective execution by detecting semantic OOD prompts using a Mahalanobis score in text-embedding space, filtering unstable generations via a directional sensitivity discrepancy metric, and enforcing final hard kinematic constraints such as joint and velocity limits before passing the generated trajectory to a low-level motion tracking controller. Extensive experiments on the Unitree G1 demonstrate that \textbf{SafeFlow} outperforms diffusion- and retargeting-based baselines in success rate, physical compliance, and inference speed while preserving motion diversity, with consistent gains across three downstream tracking controllers.
\end{abstract}

\section{Introduction}
Recent advances in text-driven motion generation~\cite{tevet2023mdm,petrovich2023tmr,chen2023mld,zhang2023t2mgpt,zhang2022motiondiffuse} have enabled humanoid robots to synthesize diverse and expressive behaviors from natural language. Beyond offline text-to-motion~\cite{serifi2024robotmdm,zhuang2025humanoidr0}, recent work has progressed toward real-time interactive control, where robots respond to streaming text commands. In particular, systems such as TextOp~\cite{xie2026textop} demonstrate a new control paradigm in which natural language serves as a continuously revisable control signal rather than a one-shot task specification, suggesting a promising direction toward intuitive, text-based humanoid control.

\begin{figure}[t]
    \centering
    \includegraphics[width=\columnwidth]{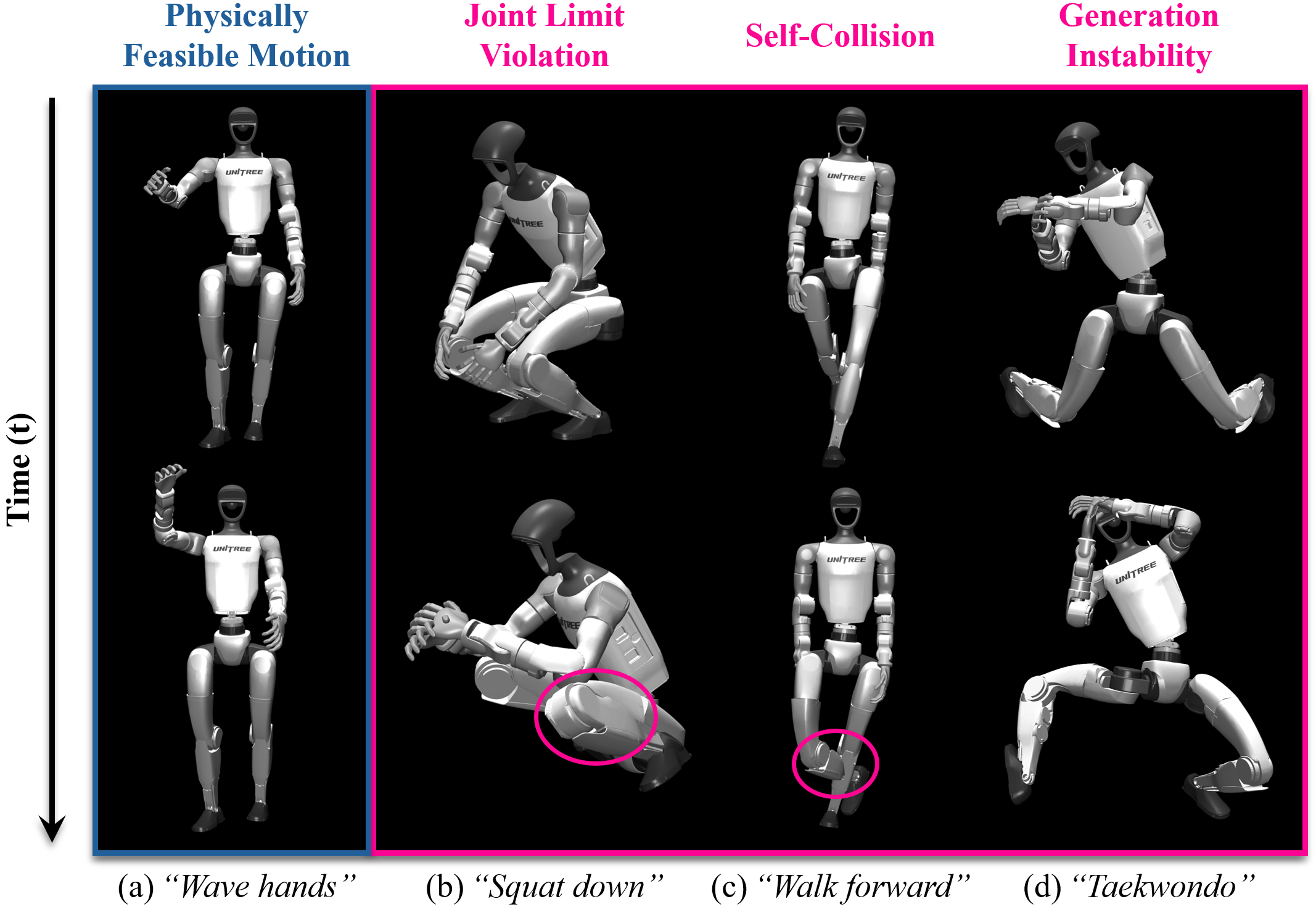}
    \caption{\textbf{Failure Cases of a Kinematics-Only Generator~\cite{xie2026textop}.} Feasible motion for a simple prompt (\textbf{a}); joint limit violations (\textbf{b}) and self-collisions (\textbf{c}) under \textit{in-distribution} commands; structural collapse under \textit{OOD} prompts (\textbf{d})---motivating physics-guided generation and runtime safety gating.}
    \label{fig:failurecases}
\end{figure}

Despite this progress, high-level motion generators often fail to produce motions that are physically executable and safe on real hardware. Kinematics-only generators~\cite{tevet2023mdm,chen2023mld,zhang2023t2mgpt} can exhibit physical hallucinations---including joint limit violations, self-collisions, and unstable balance---and although downstream motion tracking controllers can partially compensate, large violations degrade motion fidelity and can lead to unstable or unsafe behaviors.
These failures become more severe under open-ended or out-of-distribution (OOD) inputs, where generators may produce severely distorted motions unsuitable for direct execution (Fig.~\ref{fig:failurecases}). This calls for improving physical feasibility at the generation stage \emph{and} detecting and rejecting unsafe behaviors prior to execution.

To this end, we propose \textbf{SafeFlow}, a real-time text-driven humanoid whole-body control framework that combines \textit{physics-guided motion generation} with \textit{deployment-time selective execution} to improve robustness under open-ended or OOD text inputs. At the core of \textbf{SafeFlow} is a physics-guided motion generator based on rectified flow matching in a VAE latent space. Unlike purely kinematic generation~\cite{tevet2023mdm,petrovich2023tmr,chen2023mld,zhang2023t2mgpt,zhang2022motiondiffuse}, our approach incorporates physics-aware objectives relevant to real-robot execution, including joint feasibility, self-collision avoidance, stability, and motion smoothness, to steer sampling toward executable motion regions. While physics-guided sampling has been explored in character animation and offline motion generation~\cite{yuan2023physdiff}, its use for improving real-robot executability in real-time text-driven control remains underexplored. To enable real-time deployment, we further leverage reflow~\cite{liu2022reflow} distillation so that the model internalizes the physics-aware guidance, drastically reducing the required number of function evaluations while retaining physically executable behaviors. These generation-level improvements substantially enhance executability and efficiency, while deployment-time safety under ambiguous or adversarial prompts remains a complementary challenge, motivating an additional selective execution mechanism.

\textbf{SafeFlow} therefore incorporates a training-free 3-Stage Safety Gate driven by explicit risk indicators that operates hierarchically across input semantics, generation reliability, and final kinematic feasibility. We first detect semantic OOD prompts in text-embedding space, then filter structurally unstable generations by measuring directional flow sensitivity, and finally enforce a last-line kinematic screen to strictly reject motions that violate hardware constraints, including joint and velocity limits, before execution. This hierarchical filtering enables the system to proactively reject unsafe motions rather than attempting to execute all generated outputs.

By integrating physics-aware generation with indicator-driven selective execution, \textbf{SafeFlow} advances real-time text-driven humanoid control toward safe and robust deployment under unconstrained text inputs. We validate the proposed framework through extensive experiments on the Unitree G1 humanoid.
Results show that \textbf{SafeFlow} improves success rate, physical compliance, and inference speed compared to diffusion-based baselines while preserving motion diversity. Our main contributions are summarized as follows:
    \begin{itemize}
        \item
        We propose \textbf{SafeFlow}, a real-time text-driven humanoid whole-body control framework that couples physics-guided generation with deployment-time selective execution for robustness under unconstrained prompts.
        \item
        \vspace{1mm}
        We introduce \textbf{physics-guided rectified flow matching} in a VAE latent space and leverage \textbf{reflow distillation} to achieve real-time execution while significantly improving the physical feasibility and real-robot executability of generated motions.
        \item
        \vspace{1mm}
        We propose a \textbf{training-free 3-Stage Safety Gate} to proactively block unsafe behaviors under OOD prompts, utilizing explicit risk indicators: Mahalanobis semantic OOD filtering, directional sensitivity discrepancy metric for generation instability, and hard kinematic screening.
        \vspace{1mm}
    \end{itemize}

\vspace{-1mm}
\section{Related Work}
\subsection{Interactive Language-Driven Humanoid Control}
Conventional humanoid whole-body control executes task-specific commands~\cite{xue2025hugwbc,xie2025dbhl} or tracks predefined references from motion capture~\cite{mahmood2019amass} or videos~\cite{allshire2025videomimic,shen2024gvhmr} with RL-based tracking controllers~\cite{chen2025gmt,liao2025beyondmimic,zhang2025any2track,luo2025sonic}; teleoperation~\cite{he2024h2o,he2024omnih2o,ze2025twist} adds flexibility at the cost of human involvement. Generative motion models~\cite{tevet2023mdm,petrovich2023tmr,chen2023mld,zhang2023t2mgpt,zhang2022motiondiffuse} translate language into robot motion, but mostly offline.
More recently, TextOp~\cite{xie2026textop} demonstrates real-time interactive control with an autoregressive diffusion model~\cite{zhao2025dartcontrol}, allowing on-the-fly intent revision. However, such systems primarily optimize semantic alignment and can produce references that violate actuation limits or exhibit self-collisions and unstable balance---especially under out-of-distribution (OOD) commands---burdening downstream trackers~\cite{chen2025gmt,zhang2025any2track,liao2025beyondmimic}. In the same streaming setting, \textbf{SafeFlow} couples physics-aware guidance with deployment-time selective execution to improve both reference executability and robustness to unsafe OOD inputs.

\subsection{Physics-Aware Humanoid Motion Generation}
Text-conditioned motion generators~\cite{tevet2023mdm,petrovich2023tmr,chen2023mld,zhang2023t2mgpt,zhang2022motiondiffuse} often produce kinematically plausible yet physically invalid motions (\eg, foot sliding, ground penetration). To improve realism, simulator-in-the-loop diffusion methods such as PhysDiff~\cite{yuan2023physdiff} incorporate physics during sampling but introduce substantial latency. Meanwhile, physics-based RL approaches such as PhysHOI~\cite{wang2023physhoi} are tailored to virtual characters rather than hardware-constrained humanoid robots. In robotics, RobotMDM~\cite{serifi2024robotmdm} and Humanoid-R0~\cite{zhuang2025humanoidr0} bridge the kinematic-execution gap but are inherently limited to offline generation. Specifically, RobotMDM synthesizes full reference sequences from discrete prompts, while Humanoid-R0 relies on computationally heavy autoregressive generation. Both require motions to be pre-computed, making real-time streaming control infeasible. In contrast, \textbf{SafeFlow} combines physics-guided rectified flow matching with reflow distillation to generate physically executable motions at the latency required for real-time streaming control.

\subsection{Deployment-Time Safety Gating and OOD Robustness}
Real-time, interactive text-driven control exposes robots to open-ended and OOD prompts, which can induce \textit{unsafe} reference trajectories at deployment time. Most existing frameworks lack an explicit runtime mechanism to reject such unsafe references; for instance, TextOp~\cite{xie2026textop} relies on downstream tracking controllers to cope with the resulting references~\cite{liao2025beyondmimic,zhang2025any2track,chen2025gmt}, while end-to-end language policies such as LangWBC~\cite{shao2025langwbc} do not expose an explicit intermediate kinematic reference for pre-execution screening. \textbf{SafeFlow} addresses this gap with a 3-Stage Safety Gate that hierarchically screens semantic OOD inputs, generation instability via a directional sensitivity discrepancy metric, and hard kinematic limits, ensuring that only executable and safe reference trajectories are passed to the motion tracking controller.

\begin{figure*}[t]
    \centering
    \includegraphics[width=\linewidth]{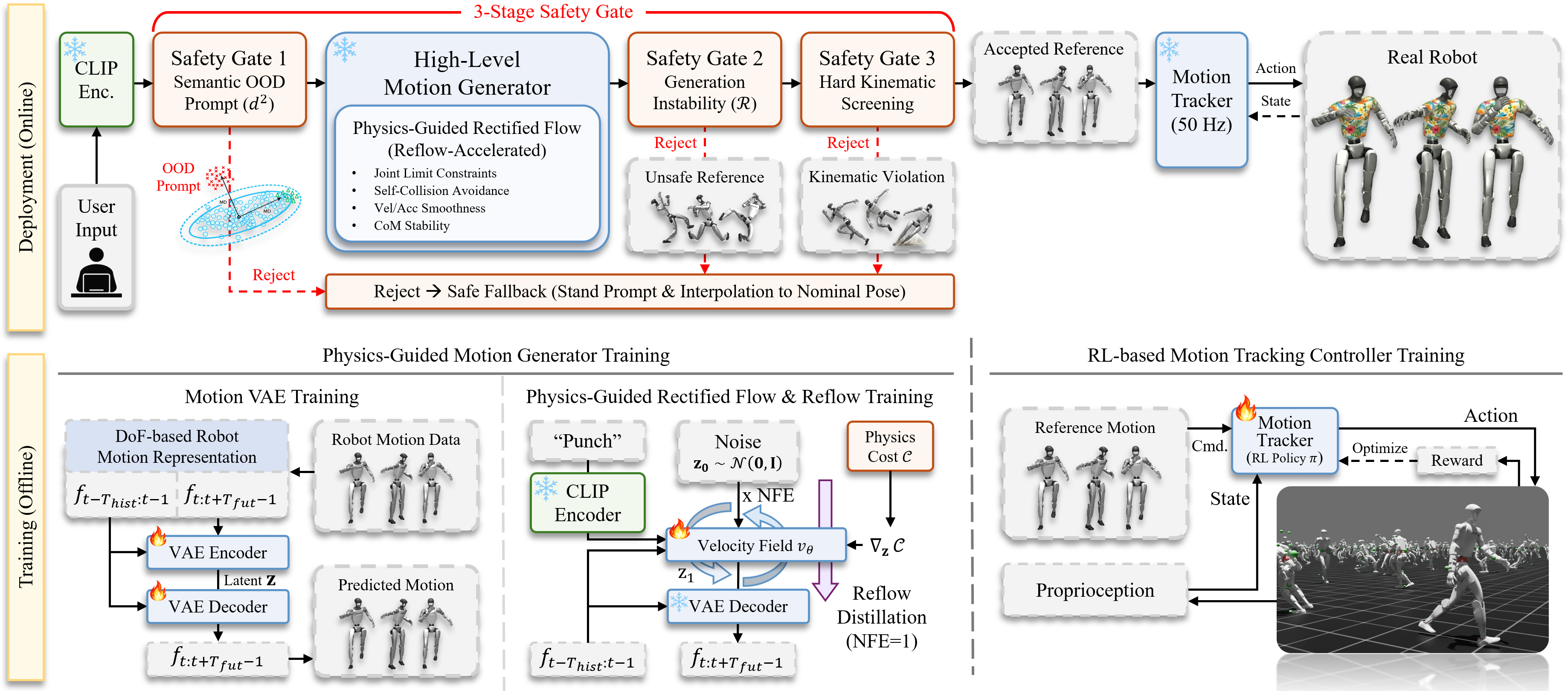}
    \caption{\textbf{Overview of SafeFlow.} \textbf{Top (Deployment, Online):} A 3-Stage Safety Gate hierarchically filters OOD semantics, generation instability, and kinematic violations. A reflow-accelerated high-level motion generator provides physically feasible reference motions. If accepted, these are executed by the downstream motion tracking controller; otherwise, a safe fallback is triggered. \textbf{Bottom (Training, Offline):} The motion generator is trained via VAE latent learning and physics-guided flow matching with reflow distillation (NFE=1). The motion tracking controller is trained in simulation via RL.}
    \label{fig:method}
\end{figure*}

\section{Method}
\subsection{Overview of SafeFlow}
\label{sec:overview}
\textbf{SafeFlow} is a two-level framework for real-time interactive text-driven humanoid control (Fig.~\ref{fig:method}), targeting two failure modes: \textit{physically infeasible} references from kinematics-only generation and \textit{unsafe} generations under open-ended or OOD prompts. To address these complementary challenges, \textbf{SafeFlow} combines physics-aware motion generation (Sec.~\ref{sec:physics_guided}) with deployment-time selective execution via a training-free 3-Stage Safety Gate (Sec.~\ref{sec:gating}), followed by an RL-based motion tracking controller (Sec.~\ref{sec:tracker}).

\noindent\textbf{Streaming Text Control.}
Following TextOp~\cite{xie2026textop}'s streaming \emph{reference generation--low-level tracking} formulation, \textbf{SafeFlow} augments the loop with deployment-time safety gating (Fig.~\ref{fig:method}). At each time step $t$, the system receives a text command $l_t$ and the robot proprioceptive state $x^{\mathrm{robot}}_{t-1}$.
We first apply the Stage-1 Safety Gate to $l_t$; if \textit{accepted}, the physics-guided motion generator $G$ produces a horizon-$T_{\mathrm{fut}}$ ($=\!8$) future reference motion sequence conditioned on the previous $T_{\mathrm{hist}}$ ($=\!2$) reference states,
\begin{equation}
x^{\mathrm{ref}}_{t:t+T_{\mathrm{fut}}-1}
= G\!\left(x^{\mathrm{ref}}_{t-T_{\mathrm{hist}}:t-1},\, l_t \right).
\end{equation}
The generated reference is then screened by the Stage-2/3 Safety Gates before execution.
The low-level tracking controller $\pi$ runs at the control rate and converts the \textit{accepted} kinematic reference into executable joint commands as
$a_\tau = \pi(x^{\mathrm{robot}}_{\tau-1}, a_{\tau-1}, x^{\mathrm{ref}}_{\tau:\tau+T_{\mathrm{ref}}-1})$,
where $x^{\mathrm{ref}}_{\tau:\tau+T_{\mathrm{ref}}-1}$ denotes the corresponding segment of the latest accepted reference.
If a prompt or generated segment is \textit{rejected}, the system executes a safe fallback motion and continues with updated streaming commands.

\subsection{Physics-Guided Rectified Flow Motion Generation}
\label{sec:physics_guided}
Real-time interactive control requires generating physically executable reference motions with low latency. We formulate the high-level motion generator $G$ using \textit{rectified flow matching} for stable kinematic modeling. However, purely kinematic generation often violates physical constraints critical for robot execution. To ensure physical feasibility, \textbf{SafeFlow} introduces \textit{physics-guided sampling} to steer generation toward executable motion regions. Finally, to support low-latency streaming, we apply \textit{reflow} distillation~\cite{liu2022reflow}, enabling highly efficient sampling via straight flow trajectories.

\vspace{1.7mm}
\noindent\textbf{Latent-Space Motion Representation.}
We represent reference motion $x^{\mathrm{ref}}_{t}$ using \textit{the same} DoF-based local incremental per-frame feature $f_t\in\mathbb{R}^{d_{\text{feat}}}$ as in TextOp~\cite{xie2026textop}. We train a VAE to learn a compact motion latent space. The encoder infers a motion latent from both past and future reference motion as $\mathbf{z}\!\sim\!\mathrm{Enc}\!\left(\cdot \mid f_{t-T_{\mathrm{hist}}:t+T_{\mathrm{fut}}-1}\right)$, while the decoder reconstructs the future reference using only the past reference and the motion latent as $f_{t:t+T_{\mathrm{fut}}-1}\!=\!\mathrm{Dec}\left(f_{t-T_{\mathrm{hist}}:t-1}, \mathbf{z}\right)$.

\vspace{1.7mm}
\noindent\textbf{Text-Conditioned Rectified Flow Matching.}
We model the text-conditional distribution of future motion latent using rectified flow matching~\cite{liu2022reflow}.
We embed the streaming text command using a CLIP text encoder~\cite{radford2021clip} as $\mathbf{e}_t\!=\!\text{CLIP}(l_t)$ and condition on the motion history $f_{t-T_{\mathrm{hist}}:t-1}$.
We learn a velocity field $v_\theta(\mathbf{z}, u \!\mid\! f_{t-T_{\mathrm{hist}}:t-1}, \mathbf{e}_t)$ that defines an Ordinary Differential Equation (ODE) transporting a noise distribution to the data distribution:
\begin{equation}
    \frac{d\mathbf{z}_u}{du} = v_\theta(\mathbf{z}_u, u \mid f_{t-T_{\mathrm{hist}}:t-1}, \mathbf{e}_t), \quad u \in [0,1].
\end{equation}
During training, we sample a ground truth motion latent $\mathbf{z}_1\!=\!\mathrm{Enc}(f_{t-T_{\mathrm{hist}}:t+T_{\mathrm{fut}}-1})$ and a noise latent $\mathbf{z}_0 \!\sim\! \mathcal{N}(\mathbf{0}, \mathbf{I})$.
We define a linear interpolation path $\mathbf{z}_u \!=\! u\mathbf{z}_1 \!+\! (1\!-\!u)\mathbf{z}_0$, which implies a constant target velocity $\mathbf{z}_1 \!-\! \mathbf{z}_0$.
The model is trained to minimize the rectified flow matching objective:
\small
\begin{equation}
    \mathcal{L}_{\text{RFM}}(\theta) = \mathbb{E} \left[ \left\| v_\theta(\mathbf{z}_u,u \!\mid\! f_{t-T_{\mathrm{hist}}:t-1}, \mathbf{e}_t) \!-\! (\mathbf{z}_1 \!-\! \mathbf{z}_0) \right\|_2^2 \right].
\end{equation}
\normalsize

At inference, we sample $\mathbf{z}_0 \sim \mathcal{N}(\mathbf{0}, \mathbf{I})$ and integrate the ODE from $u\!=\!0$ to $u\!=\!1$ using an explicit solver (\eg, Euler) with $N$ steps (\ie, NFE=$N$) to obtain the generated motion latent $\mathbf{z}_1$, then decode $f_{t:t+T_{\mathrm{fut}}-1}\!=\!\mathrm{Dec}(f_{t-T_{\mathrm{hist}}:t-1}, \mathbf{z}_1)$.

\vspace{1.7mm}
\noindent\textbf{Physics-Guided Sampling.}
Text conditioning does not guarantee physical feasibility on a real robot, so \textbf{SafeFlow} employs \textit{physics-guided sampling} to steer the motion latent toward executable manifolds, imposing constraints purely at inference time without retraining.
Let $f_{t:t+T_{\mathrm{fut}}-1}=\mathrm{Dec}(f_{t-T_{\mathrm{hist}}:t-1},\mathbf{z}_u)$ denote the decoded motion feature sequence at ODE integration time $u$, with $x^{\mathrm{ref}}_{t:t+T_{\mathrm{fut}}-1}$ denoting the corresponding kinematic reference trajectory.

We define a differentiable physics cost $\mathcal{C}$ to quantify executability violations.
While gradient-based guidance has been explored in character animation and offline motion~\cite{tevet2023mdm,yuan2023physdiff,xie2024omnicontrol}, to the best of our knowledge, \textbf{SafeFlow} is the first to adapt this mechanism for \textit{real-robot executability} by enforcing strict hardware limits, self-collision avoidance, and postural stability via CoM regularization. The total cost is a weighted sum of four terms as $\mathcal{C}\!\left(x^{\mathrm{ref}}_{t:t+T_{\mathrm{fut}}-1}\right) \!=\! \sum_i \lambda_i \mathcal{C}_i$, where $\mathcal{C}_i$ represents specific constraints (\textit{detailed below}). During ODE integration, we compute the gradient of the cost with respect to the \textit{generated motion latent} $\mathbf{z}$.
This involves passing $\mathbf{z}$ through the frozen VAE decoder to reconstruct the kinematic reference for cost evaluation, and then backpropagating.
We then steer the flow using the \textit{guided velocity field} $\tilde{v}_\theta$:
\begin{equation}
    \begin{split}
    \tilde v_\theta(\mathbf{z},u)
    &=
    v_\theta(\mathbf{z},u \mid f_{t-T_{\mathrm{hist}}:t-1},\mathbf{e}_t)
    \\ &\quad - \alpha(u)\,\nabla_{\mathbf{z}}\,
    \mathcal{C}\!\left(\mathrm{Dec}(f_{t-T_{\mathrm{hist}}:t-1},\mathbf{z})\right),
    \label{eq:guided_velocity}
    \end{split}
    \end{equation}
where $\alpha(u)$ is a time-dependent guidance scale. We use this steered velocity $\tilde v_\theta$ for numerical integration to push the trajectory toward physically feasible regions.

We define $\mathcal{C}$ as a weighted sum of four terms designed for real-robot executability. Let $\mathbf{q}_{\tau}$ be the joint configuration at time $\tau$ decoded from $f_\tau$.

\vspace{1.7mm}
\noindent\textit{(1) Joint Limit \& (2) Self-Collision:}
To strictly enforce hardware limits and prevent physical penetrations, we penalize violations using ReLU-squared barriers:
\small\begin{equation}
    \begin{split}
        \mathcal{C}_{\mathrm{lim}} &= \sum_{\tau, j} \Big(\mathrm{ReLU}(q_{\tau,j}-q_j^{\max})^2 + \mathrm{ReLU}(q_j^{\min}-q_{\tau,j})^2\Big), \\
        \mathcal{C}_{\mathrm{col}} &= \sum_{\tau, (a,b)\in\mathcal{P}}
        \mathrm{ReLU}\!\big((r_a + r_b + m) - d_{ab}(\mathbf{q}_\tau)\big)^2,
    \end{split}
\end{equation}\normalsize
where $r_a, r_b$ are collision sphere radii for links $a, b$, $d_{ab}$ is their Euclidean distance, and $m$ is a safety margin.

\vspace{1.7mm}
\noindent\textit{(3) Smoothness \& (4) CoM Stability:}
To generate smooth, jitter-free motions suitable for tracking, we regularize high-order derivatives of joints and the Center of Mass (CoM).
Let $\mathbf{c}(\mathbf{q}) = \frac{\sum m_i \mathbf{p}_i(\mathbf{q})}{\sum m_i}$ be the global CoM position computed via forward kinematics, where $m_i$ and $\mathbf{p}_i$ are the mass and position of the $i$-th link.
\small\begin{equation}
    \begin{split}
        \mathcal{C}_{\mathrm{sm}} &= \sum_{\tau} \left( \|\dot{\mathbf{q}}_\tau\|^2 + \beta_q \|\ddot{\mathbf{q}}_\tau\|^2 \right), \\
        \mathcal{C}_{\mathrm{stab}} &= \sum_{\tau} \left( \|\dot{\mathbf{c}}(\mathbf{q}_\tau)\|^2 + \beta_c \|\ddot{\mathbf{c}}(\mathbf{q}_\tau)\|^2 \right).
    \end{split}
\end{equation}\normalsize
Here, time derivatives are computed via finite differences.

\vspace{1.7mm}
\noindent\textbf{Physics-Aware Reflow.}
Physics-guided sampling improves executability but increases latency due to iterative gradient computations ($\nabla_{\mathbf{z}}\mathcal{C}$).
To enable real-time control, we apply the \textit{reflow} procedure~\cite{liu2022reflow} to distill the guided trajectories into a straighter velocity field.
We generate synthetic pairs $(\mathbf{z}_0, \mathbf{z}_1^{\text{guided}})$, where $\mathbf{z}_1^{\text{guided}}$ is obtained through the expensive guided integration in Eq.~\ref{eq:guided_velocity}.
We then retrain the model along the straight paths connecting $\mathbf{z}_0$ and $\mathbf{z}_1^{\text{guided}}$.
This process \textit{internalizes} the physics-aware guidance into the network weights, eliminating online cost-gradient computation and enabling physically executable motion generation with fewer function evaluations (\eg, NFE=1) at deployment.

\subsection{Selective Execution via 3-Stage Safety Gate}
\label{sec:gating}
While physics-guided motion generation improves average executability, it cannot inherently prevent failures caused by open-ended or OOD text inputs. Such inputs often reside in sparse regions of the training distribution, leading to physical hallucinations or structurally unstable motions. To ensure robust deployment without compromising real-time interactivity, \textbf{SafeFlow} introduces a \textit{training-free} selective execution mechanism. As a hierarchical \textit{firewall}, it filters failure modes at the \textit{input semantic}, \textit{latent generative}, and \textit{output kinematic} levels, rejecting unsafe references with acceptable latency before reaching the motion tracking controller.

\vspace{1.7mm}
\noindent\textbf{Stage 1: Semantic OOD Filtering (Input Level).} Standard generators often fail unpredictably when facing out-of-distribution (OOD) prompts. We detect these efficiently in the CLIP~\cite{radford2021clip} text embedding space.
Since the statistics of training prompts (mean $\boldsymbol{\mu}$ and covariance $\boldsymbol{\Sigma}$) are \textit{pre-computed offline}, inference requires only a lightweight Mahalanobis distance calculation on the streaming text embedding $\mathbf{e}_t$: $d^2(\mathbf{e}_t)=(\mathbf{e}_t-\boldsymbol{\mu})^\top \boldsymbol{\Sigma}^{-1}(\mathbf{e}_t-\boldsymbol{\mu})$.
The threshold $\tau_{\text{sem}}$ is calibrated to the $N$-th percentile of distances computed on the training set. Prompts satisfying $d^2 \!>\! \tau_{\text{sem}}$ are rejected instantly, bypassing the motion generator to prevent synthesizing \textit{undefined} reference motions.

\vspace{1.7mm}
\noindent\textbf{Stage 2: Generation Instability Filtering (Model Level).}
Even for valid prompts, the flow can traverse unstable regions where the vector field becomes highly anisotropic.
To detect this structural instability, we introduce a novel \textit{directional sensitivity discrepancy} metric.
The key intuition is that, in stable regions, the flow responds consistently to perturbations across directions, whereas high variance indicates fragility to specific directional noise.

We estimate this by probing the Jacobian $J = \partial v_\theta / \partial \mathbf{z}$ along $M$ random unit vectors $\{\boldsymbol{\epsilon}_m\}_{m=1}^{M}$ (\eg, $M\!=\!16$).
First, we compute the directional sensitivity scalar $g_m$ for each probe using a finite-difference approximation:
\begin{equation}
    g_m \approx \boldsymbol{\epsilon}_m^\top \left( \frac{v_\theta(\mathbf{z}+\delta\boldsymbol{\epsilon}_m) - v_\theta(\mathbf{z})}{\delta} \right) \approx \boldsymbol{\epsilon}_m^\top J \boldsymbol{\epsilon}_m,
\end{equation}
where $\delta$ is a small perturbation. This scalar $g_m$ represents the expansion or contraction of the flow along $\boldsymbol{\epsilon}_m$.
Finally, we define the \textit{generation instability score} $\mathcal{R}$ as the standard deviation of these sensitivities,
{\small $\mathcal{R}\!=\!\sqrt{\frac{1}{M}\sum_{m=1}^{M} (g_m\!-\!\bar{g})^2}$},
where $\bar{g}$ denotes the mean of $\{g_m\}$.
Leveraging \textit{parallel batching}, this computation incurs negligible latency, enabling \textit{real-time risk monitoring} during generation.
A high $\mathcal{R} (> \tau_{\text{stab}})$ indicates that the flow field is disjointed or near-singular, triggering early rejection to prevent executing \textit{structurally unreliable} reference motions.

\vspace{1.7mm}
\noindent\textbf{Stage 3: Hard Kinematic Screening (Output Level).}
As a final fail-safe, we perform a lightweight, deterministic screen on $x^{\mathrm{ref}}$, strictly rejecting segments that violate joint position bounds ($q_{j} \notin [q^{\min}_j, q^{\max}_j]$) or dynamic constraints ($|\dot{q}_{j}| > \dot{q}^{\max}_j$, $|\ddot{q}_{j}| > \ddot{q}^{\max}_j$).
While this local check cannot guarantee global stability (\eg, balance), it serves as a necessary \textit{last-line defense} to prevent immediate actuator damage.
If a rejection is triggered at any stage, the system executes a safe fallback by replacing the current user command with a ``stand'' prompt while simultaneously interpolating to a nominal pose, and awaits the next command.

\subsection{RL-Based Motion Tracking Controller}
\label{sec:tracker}
We adopt a goal-conditioned RL motion tracking controller trained with PPO in Isaac Lab~\cite{mittal2025isaaclab}.
The controller outputs residual joint corrections $\Delta q_\pi$, forming control targets as $q_{\text{target}}\!=\!q_{\text{ref}}(t)\!+\!\Delta q_\pi$, which improves robustness to imperfect references. To enhance generalization, future reference observations are expressed in the body-local frame (linear/angular velocities, root height, roll--pitch, and joint targets), ensuring invariance to global position and heading.

\subsection{Implementation Details}
\label{sec:impl}
\noindent\textbf{Physics-Guided Motion Generator \& Safety Gating.}
Our model is trained on BABEL~\cite{Punnakkal2021babel} retargeted to Unitree G1.
We follow TextOp~\cite{xie2026textop} for data preprocessing and splits.
Training tuples use sliding windows of $(T_{\mathrm{hist}},T_{\mathrm{fut}})\!=\!(2,8)$, enabling the generator to operate at 6.25\,Hz.
Building upon TextOp, we adopt its exact motion representations, Transformer architectures, and remaining training hyperparameters, training the velocity field $v_\theta$ for 200k iterations.
We build a physics-guided teacher ({\small \textbf{SafeFlow}\,(+\,Guid.)}, NFE\,=\,10) using classifier-free guidance (CFG) decaying from 5.0 to 3.0.
Physics-guided sampling uses $\lambda_{\mathrm{lim}}\!=\!\lambda_{\mathrm{stab}}\!=\!1.0$, $\!\lambda_{\mathrm{sm}}\!=\!0.1$, $\lambda_{\mathrm{col}}\!=\!0.01$, $\beta_q\!=\!50.0$, and $\beta_c\!=\!10.0$.
We apply physics guidance scale $\alpha(u)$ with a linearly increasing schedule from $500$ to $10{,}000$ across the denoising trajectory, with per-element gradient clamping at $\pm 0.2$.
For real-time deployment ({\small \textbf{SafeFlow}\,(+\,Guid.\,\&\,Reflow)}, NFE\,=\,1), we distill guided ODE trajectories into straight paths via reflow for an additional 200k iterations.
Self-collision is computed over 14 link pairs with sphere radii $r \in [0.03, 0.10]\,\mathrm{m}$ and margin $m = 0.03\,\mathrm{m}$.
For safety gating, Stage~1 uses $\tau_{\mathrm{sem}}$ calibrated to accept 95\% of training prompts (trade-off analyzed in Table~\ref{tab:ood_filtering}).
Stage~2 evaluates $\mathcal{R}$ using 16 probes ($\delta\!=\!10^{-6}$) with threshold $\tau_{\mathrm{stab}}\!=\!5.0$, which matches the F1-optimal value on evaluated rollouts ($\tau^*\!=\!4.9$; Sec.~\ref{sec:exp_safety}).
Stage~3 enforces G1 hardware limits.

\vspace{1.7mm}
\noindent\textbf{Motion Tracking Controller.}
Since our contributions lie in the physics-guided generator and the 3-Stage Safety Gate, we adopt the same RL tracking setup as TextOp~\cite{xie2026textop} (\ie, training data, observations, rewards, and domain randomization). For the primary comparison, the identical trained tracking policy, referred to as the \emph{Base Tracker}~\cite{xie2026textop}, is used for all evaluated generation methods. The controller runs at 50\,Hz on the onboard Jetson Orin. Cross-tracker generalization is further evaluated with SONIC~\cite{luo2025sonic} and MOSAIC~\cite{jiang2026mosaic} in Table~\ref{tab:executability}.

\section{Experiments}
\label{sec:experiments}

\begin{table*}[t]
    \centering
    \renewcommand{\arraystretch}{1.05}
    \setlength{\tabcolsep}{6pt}

    \caption{
    \textbf{Physical Executability and Cross-Tracker Tracking Fidelity.}
    JV/SC denote joint-limit-violation/self-collision rates.
    Tracking errors are evaluated on valid frames before failure (common-prefix),
    with success-only values in parentheses.
    All tracking errors are computed relative to each generator's own reference.
    The Base Tracker is shared with TextOp~\cite{xie2026textop};
    SONIC~\cite{luo2025sonic} and MOSAIC~\cite{jiang2026mosaic} use their unmodified deployment stacks.
    $^{\dagger}$JV is measured post-retargeting (IK clamps limits).
    }
    \vspace{-1mm}
    \label{tab:executability}

    \footnotesize
    \resizebox{\textwidth}{!}{%
    \begin{tabular}{@{}lccccccccccc@{}}
        \toprule
        &
        \multicolumn{2}{c}{\textbf{Generator-Only}}
        &
        \multicolumn{9}{c}{\textbf{System-Level Tracking Fidelity}}
        \\

        \cmidrule(lr){2-3}
        \cmidrule(lr){4-12}

        ~\textbf{Method}
        &
        \textbf{JV$\downarrow$}
        &
        \textbf{SC$\downarrow$}
        &
        \multicolumn{3}{c}{\textbf{Base Tracker}~\cite{xie2026textop}}
        &
        \multicolumn{3}{c}{\textbf{SONIC}~\cite{luo2025sonic}}
        &
        \multicolumn{3}{c}{\textbf{MOSAIC}~\cite{jiang2026mosaic}}
        \\

        \cmidrule(lr){4-6}
        \cmidrule(lr){7-9}
        \cmidrule(lr){10-12}

        & & &
        \textbf{Succ.$\uparrow$}
        & $\mathbf{E}_{\mathrm{mpjpe}}\downarrow$
        & $\mathbf{E}_{\mathrm{dof}}\downarrow$
        & \textbf{Succ.$\uparrow$}
        & $\mathbf{E}_{\mathrm{mpjpe}}\downarrow$
        & $\mathbf{E}_{\mathrm{dof}}\downarrow$
        & \textbf{Succ.$\uparrow$}
        & $\mathbf{E}_{\mathrm{mpjpe}}\downarrow$
        & $\mathbf{E}_{\mathrm{dof}}\downarrow$
        ~\\

        \midrule

        ~MDM+GMR~\cite{tevet2023mdm} {\footnotesize(offline)}
        & 0.79\%$^{\dagger}$
        & 30.48\%
        & 87.3\%
        & \underline{45.2}\,(42.2)
        & 4.96\,(4.74)
        & 92.6\%
        & 40.3\,(38.6)
        & 8.97\,(8.71)
        & 89.0\%
        & 44.3\,(42.1)
        & 7.53\,(7.23)
        ~\\

        ~TextOp~\cite{xie2026textop}
        & 43.14\%
        & 11.05\%
        & 80.6\%
        & 82.5\,(79.2)
        & 5.94\,(5.72)
        & 81.3\%
        & 45.9\,(43.1)
        & 8.36\,(8.05)
        & 74.6\%
        & 49.3\,(46.3)
        & 8.38\,(8.05)
        ~\\

        ~+critic-align (RobotMDM-style)~\cite{serifi2024robotmdm}
        & 24.08\%
        & 12.94\%
        & 87.6\%
        & 92.2\,(89.5)
        & 5.28\,(5.11)
        & 84.5\%
        & 43.7\,(41.3)
        & 8.19\,(7.90)
        & 79.3\%
        & 47.1\,(43.8)
        & 7.68\,(7.30)
        ~\\

        \midrule

        \rowcolor{black!8}
        ~\textbf{SafeFlow}\,(Flow)
        & 12.75\%
        & 7.25\%
        & 92.8\%
        & 59.6\,(58.3)
        & 4.73\,(4.64)
        & 91.9\%
        & 34.4\,(33.2)
        & 6.42\,(6.26)
        & 90.2\%
        & 36.2\,(35.1)
        & 6.13\,(6.00)
        ~\\

        \rowcolor{black!8}
        ~\textbf{SafeFlow}\,(+\,Guid.)
        & \underline{6.32}\%
        & \underline{4.39}\%
        & \underline{97.9}\%
        & 51.2\,(50.9)
        & \underline{4.43}\,(4.41)
        & \underline{96.4}\%
        & \underline{30.8}\,(30.3)
        & \underline{5.77}\,(5.71)
        & \underline{97.1}\%
        & \underline{31.2}\,(30.9)
        & \underline{5.52}\,(5.48)
        ~\\

        \rowcolor{black!8}
        ~\textbf{SafeFlow}\,(+\,Guid.\,\&\,Reflow)
        & \textbf{3.08}\%
        & \textbf{1.42}\%
        & \textbf{98.6}\%
        & \textbf{44.0}\,(43.9)
        & \textbf{4.32}\,(4.30)
        & \textbf{99.1}\%
        & \textbf{29.6}\,(29.5)
        & \textbf{5.15}\,(5.14)
        & \textbf{98.7}\%
        & \textbf{27.6}\,(27.4)
        & \textbf{5.00}\,(4.99)
        ~\\

        \bottomrule
    \end{tabular}}
\end{table*}

We evaluate the effectiveness of \textbf{SafeFlow} through a combination of extensive simulation studies and real-world deployment on the Unitree G1 humanoid. Our experiments are designed to validate the system's physical executability, deployment-time safety and robustness, computational efficiency, and overall practical performance.
Specifically, the evaluation aims to answer the following questions:
\begin{itemize}
    \item \textbf{Q1 (Executability):} How much does physics-guided generation improve physical feasibility of reference motions and downstream tracking fidelity across trackers?
    \item \textbf{Q2 (Safety and Robustness):} Can the 3-Stage Safety Gate detect and filter out OOD prompts and generation instability to improve deployment-time safety?
    \item \textbf{Q3 (Real-Time Performance):} Do the reflow-\hspace{0pt}accelerated motion generator and safety gating pipeline achieve the low latency required for real-time control?
    \item \textbf{Q4 (Real-Robot Deployment):} Can \textbf{SafeFlow} transfer to real hardware to enable robust interactive control under challenging and potentially unsafe commands?
\end{itemize}

\subsection{Experimental Setup}
To evaluate \textbf{SafeFlow} across \textit{Physical Executability}, \textit{Deployment-Time Safety}, and \textit{Computational Efficiency}, the motion generator is trained offline in PyTorch~\cite{paszke2019pytorch}, while the tracking controller is trained in Isaac Lab~\cite{mittal2025isaaclab}. System-level evaluations of the integrated pipeline are conducted in MuJoCo~\cite{todorov2012mujoco} to assess closed-loop Unitree G1 behavior.

\vspace{1.7mm}
\noindent\textbf{Baselines.}
We compare against (i) TextOp~\cite{xie2026textop}, the state-of-the-art autoregressive diffusion baseline; (ii) \textit{MDM\,+\,GMR}, an offline (non-streaming) kinematics-only pipeline retargeting pretrained MDM~\cite{tevet2023mdm} generations to the G1 with the same GMR retargeting used for our training data (SMPL parameters recovered from MDM's rotation channels; median $+1.5$\,cm SMPL-space FK error vs.\ per-motion SMPLify~\cite{bogo2016smplify} fits on a 1{,}344-motion subset); and (iii) \textit{TextOp\,+\,critic-align}, a RobotMDM~\cite{serifi2024robotmdm}-style baseline fine-tuning the same diffusion backbone with a tracking-outcome critic (motion-level AUC 0.987, trained on 9.4k evaluated rollouts).

\vspace{1.7mm}
\noindent\textbf{Tracking Controllers.}
To verify that our conclusions are not tied to a particular controller, all generation methods are additionally evaluated with two public large-scale trackers---SONIC~\cite{luo2025sonic} and MOSAIC~\cite{jiang2026mosaic}---using their unmodified deployment stacks in sim-to-sim under identical prompts and a common success criterion. We report Succ., $E_{\mathrm{mpjpe}}$, and $E_{\mathrm{dof}}$ for all three trackers; for comparability, $E_{\mathrm{dof}}$ is averaged over the 23 non-wrist joints shared across all stacks.

\subsection{Physical Executability (Q1)}
\label{sec:exp_executability}
We first evaluate whether the physics-guided generation improves the physical feasibility of reference motions. To strictly decouple the performance of the motion generator from the capabilities of the downstream tracking controller, we assess executability in two stages: \textit{Generator-Only} (kinematic evaluation prior to tracking) and \textit{System-Level} (closed-loop evaluation with the tracking controller). For evaluation, we utilize the BABEL~\cite{Punnakkal2021babel} validation prompts, generating 1,000 motion frames per prompt and reporting the averages.

\begin{figure}[t]
    \centering
    \includegraphics[width=\columnwidth]{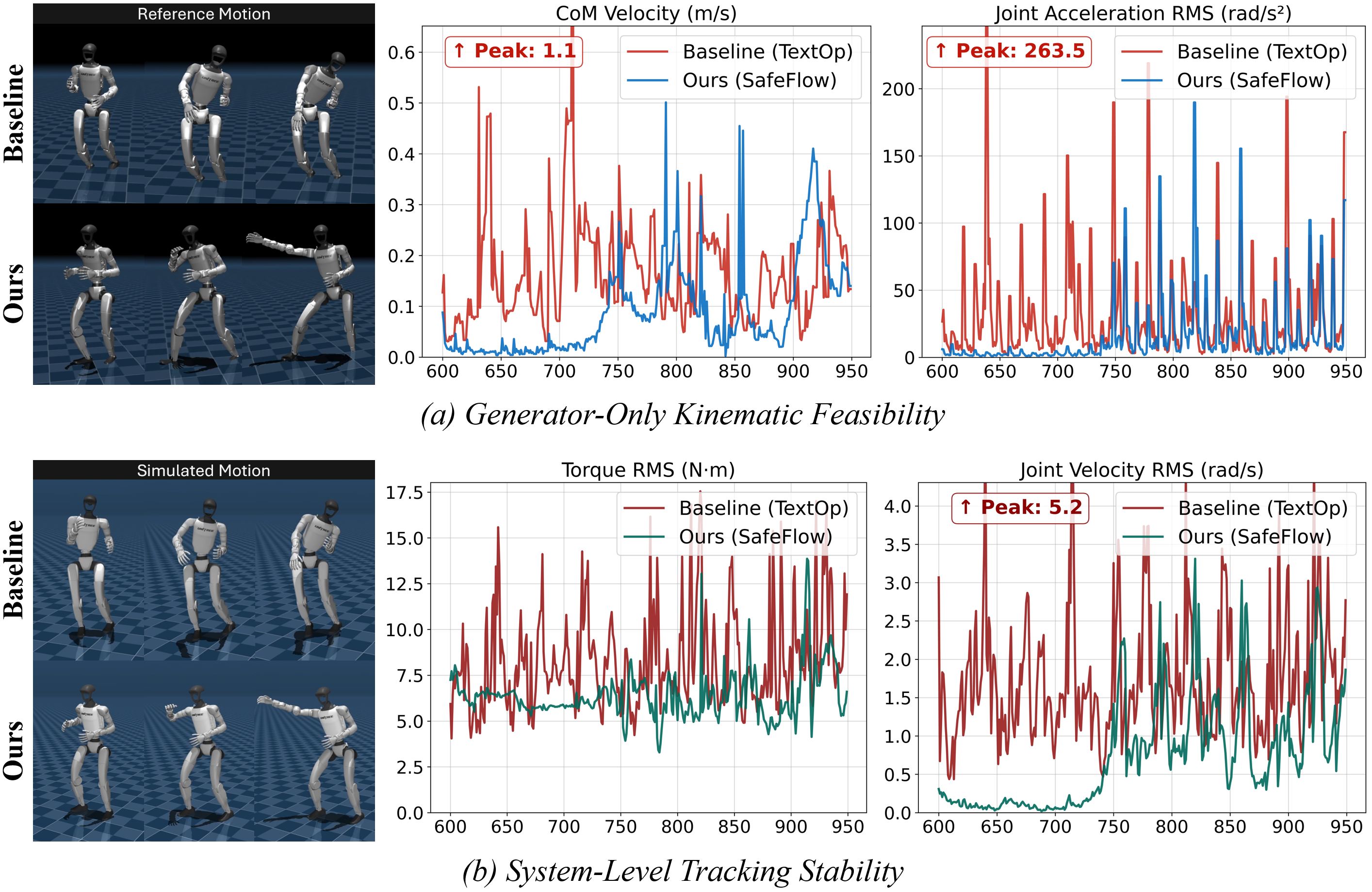}
    \vspace{-6mm}
    \caption{\textbf{Kinematic Feasibility and Tracking Stability.} Despite generating dynamic motions (left), our full pipeline, {\footnotesize \textbf{SafeFlow}\,(+\,Guid.\,\&\,Reflow)}, stabilizes kinematic references and improves tracking. (a) \textit{Generator-only}: \textbf{SafeFlow} suppresses erratic spikes in CoM velocity and joint acceleration. (b) \textit{System-level}: \textbf{SafeFlow} mitigates torque chattering and joint velocity spikes, enabling hardware-safe tracking. The x-axis represents time in frames, showing a representative active segment (frames 600--950).}
    \label{fig:physics_graphs}
\end{figure}

\vspace{1.7mm}
\noindent\textbf{Generator-Only Kinematic Feasibility.}
We measure the \textit{Joint Limit Violation Rate} (JV, \%) and \textit{Self-Collision Rate} (SC, \%) of generated motions.
As shown in Table~\ref{tab:executability}, TextOp~\cite{xie2026textop} frequently violates hardware limits (JV:\,43.14\%, SC:\,11.05\%).
Notably, adopting our base Flow Matching formulation ({\small \textbf{SafeFlow}\,(Flow)}) already reduces violations to 12.75\% JV and 7.25\% SC.
Building upon this foundation, our physics-guided sampling ({\small \textbf{SafeFlow}\,(+\,Guid.)}) further reduces these errors to 6.32\% and 4.39\%, respectively.
The reflow-distilled model ({\small \textbf{SafeFlow}\,(+\,Guid.\,\&\,Reflow)}) further lowers them to 3.08\% JV and 1.42\% SC while enabling single-step (NFE=1) generation for real-time control (Tab.~\ref{tab:latency}).
Among alternative generator baselines, \textit{MDM\,+\,GMR} exhibits a low post-retargeting JV of 0.79\% because IK clamps joint limits, but still incurs a high 30.48\% self-collision rate.
Outcome-supervised \textit{TextOp\,+\,critic-align} reduces TextOp's JV from 43.14\% to 24.08\%, but leaves SC at 12.94\%, remaining substantially less feasible than \textbf{SafeFlow}.
\textit{CoM velocity} and \textit{joint acceleration} plots (Fig.~\ref{fig:physics_graphs}\,(a)) further show that \textbf{SafeFlow} suppresses erratic spikes (baseline peak: 263.5\,$\mathrm{rad/s^2}$), yielding more trackable references.

\vspace{1.7mm}
\noindent\textbf{System-Level Tracking Fidelity.}
We next evaluate how reference feasibility translates to closed-loop tracking.
Across all trackers, we report the \textit{Success Rate} (Succ., defined as completing the sequence without falling, \ie, base height\,$>$\,0.3\,$\mathrm{m}$) and \textit{MPJPE} ({\small $E_{\mathrm{mpjpe}}$,\,$\mathrm{mm}$}).
We also report mean joint-angle error $E_{\mathrm{dof}}$ (deg), averaged over the 23 non-wrist joints shared across all trackers.
On the Base Tracker, {\small \textbf{SafeFlow}\,(+\,Guid.\,\&\,Reflow)} raises success from 80.6\% to 98.6\% and reduces $E_{\mathrm{mpjpe}}$ from 82.5 to 44.0\,$\mathrm{mm}$ vs.\ TextOp.
The kinematics-only \textit{MDM\,+\,GMR} pipeline reaches only 87.3\% success despite 3.5$\times$ lower mean joint acceleration, while \textit{TextOp\,+\,critic-align} improves success to 87.6\% but worsens $E_{\mathrm{mpjpe}}$ to 92.2\,$\mathrm{mm}$.
These results suggest that post-hoc retargeting or outcome supervision alone does not provide the same tracking benefits as physics-aware generation within the sampler.
Moreover, \textit{torque} and \textit{joint velocity} plots (Fig.~\ref{fig:physics_graphs}\,(b)) show that \textbf{SafeFlow} mitigates torque chattering and erratic velocity spikes (\eg, baseline peak: 5.2\,$\mathrm{rad/s}$).

\vspace{1.7mm}
\noindent\textbf{Tracker Generalization.}
This improvement is not specific to the Base Tracker.
Table~\ref{tab:executability} shows that the full \textbf{SafeFlow} model reaches 98.6--99.1\% success across the Base Tracker~\cite{xie2026textop}, SONIC~\cite{luo2025sonic}, and MOSAIC~\cite{jiang2026mosaic}, compared with 74.6--81.3\% for TextOp, corresponding to gains of 17.8--24.1\,pp.
It also consistently reduces $E_{\mathrm{mpjpe}}$ from 82.5/45.9/49.3\,$\mathrm{mm}$ to 44.0/29.6/27.6\,$\mathrm{mm}$ across the three trackers, respectively.
These consistent gains indicate that the benefits of physics-aware generation are not specific to any downstream tracker.

\vspace{1.7mm}
\noindent\textbf{Diversity Preservation.}
Improved physical feasibility does not sacrifice motion diversity.
We measure \textit{Multimodality} ({\small MModality})~\cite{guo2022mmodality} as the mean pairwise L2 distance in 29-DoF joint-angle space ($\mathrm{rad}$) over 10 generations per prompt.
Across all 2{,}362 prompts, TextOp shows higher diversity (1.40\,vs.\,1.09\,$\mathrm{rad}$), but this gap is largely driven by unstable motions.
Among the 1{,}889 prompts where tracking succeeds for both methods, the gap shrinks to 1.26\,vs.\,1.06\,$\mathrm{rad}$; among the 915 prompts with no joint-limit violations from either method, the two are nearly identical (1.00\,vs.\,0.99\,$\mathrm{rad}$,\,$\Delta\!=\!1.1\%$).
On the 437 prompts where only TextOp fails, its multimodality rises to 1.99\,$\mathrm{rad}$---66\% above \textbf{SafeFlow}'s 1.20\,$\mathrm{rad}$ on the same prompts---indicating that much of TextOp's apparent diversity advantage stems from physically implausible motions rather than meaningful variation.

\subsection{Deployment-Time Safety and Robustness (Q2)}
\label{sec:exp_safety}
We evaluate the 3-Stage Safety Gate's ability to filter out-of-distribution (OOD) prompts and generative instabilities, improving deployment-time safety.

\vspace{1.7mm}
\noindent\textbf{Stage~1: Input-Level Semantic Filtering.}
To evaluate robustness to unseen and potentially unsafe text commands, we generate two OOD prompt sets with an LLM~\cite{geminiteam2025gemini} (100 each): \textbf{Type A (Unseen Verbs)}, containing actions absent from training (\eg, ``levitate'', ``crochet a sweater''), and \textbf{Type B (Extreme Dynamics)}, containing extreme acrobatic motions exceeding hardware limits (\eg, ``flying tornado kick''). We compare them against 2{,}362 ID BABEL validation prompts.

We use a Mahalanobis distance-based filter with $\tau_{95}$, calibrated to pass 95\% of training prompts. As Table~\ref{tab:ood_filtering} shows, Stage~1 achieves high AUROCs (0.9872, 0.9715) for Types A and B, rejecting 91\%/77\% of OOD prompts while rejecting only 3.7\% of ID prompts. A more conservative $\tau_{90}$ raises OOD rejection to 95\%/93\% at 9.4\% ID rejection, demonstrating a tunable safety--coverage trade-off. Stage~1 serves as a semantic OOD filter rather than a physical-failure predictor; accepted prompts are therefore further screened by Stages~2--3. Stage~2 directly monitors the flow dynamics during inference to detect generative instabilities.

\begin{table}[t]
    \centering
    \renewcommand{\arraystretch}{1.08}
    \setlength{\tabcolsep}{6pt}
    \caption{\textbf{Stage 1 Semantic OOD Filtering.} With $\tau_{95}$ passing 95\% of training prompts, Stage~1 rejects OOD inputs while preserving high ID coverage.}
    \vspace{-2mm}
    \label{tab:ood_filtering}
    \resizebox{\columnwidth}{!}{
    \begin{tabular}{lccc}
        \toprule
        \textbf{Prompt Set} & \textbf{Category} & \textbf{AUROC$\uparrow$} & \textbf{Accept Rate\,@\,$\tau_{95}$} \\
        \midrule
        ID (Val.) & In-Distribution & -- & 96.3\% (2275/2362) \\
        OOD (Type A) & Unseen Verbs & 0.9872 & 9.0\% (9/100) \\
        OOD (Type B) & Extreme Dynamics & 0.9715 & 23.0\% (23/100) \\
        \bottomrule
    \end{tabular}
    }
\end{table}

\begin{figure}[t]
    \centering
    \includegraphics[width=\columnwidth]{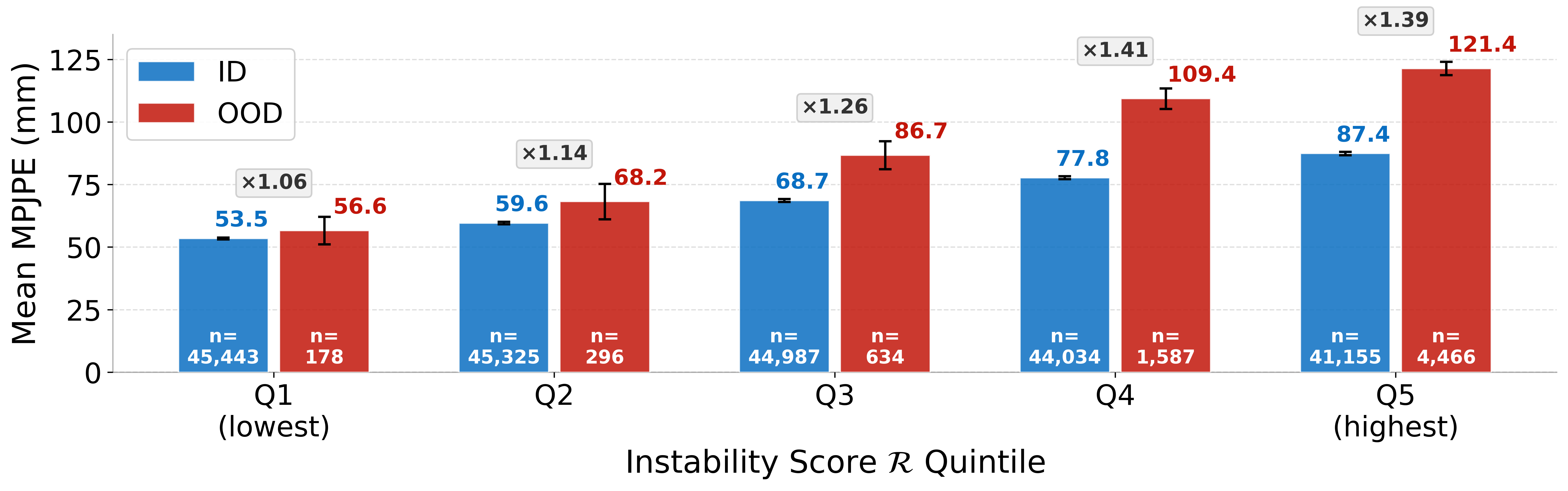}
    \vspace{-6mm}
    \caption{\textbf{$\mathcal{R}$ Detects Failure-Prone References.} Mean tracking MPJPE of 10-frame windows increases monotonically across shared absolute $\mathcal{R}$ quintiles in both ID and OOD. Even ID prompts produce high-instability windows (ID Q5, 87.4\,$\mathrm{mm}$) exceeding low-$\mathcal{R}$ OOD windows (OOD Q1, 56.6\,$\mathrm{mm}$)---showing Stage~1 alone is insufficient and motivating Stage~2.}
    \label{fig:instability_quintile}
\end{figure}

\begin{figure}[t]
    \centering
    \includegraphics[width=\columnwidth]{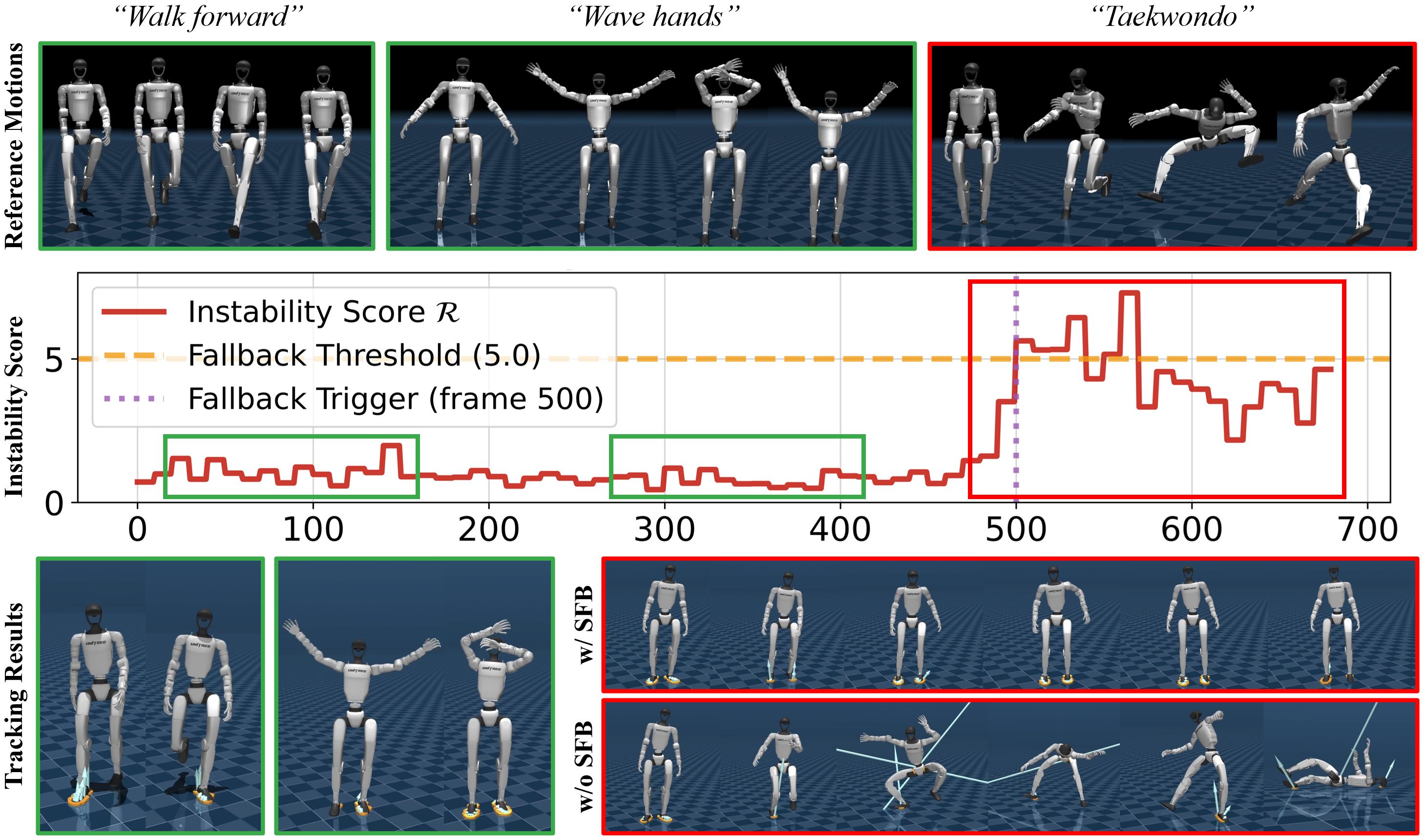}
    \vspace{-6mm}
    \caption{\textbf{Instability-Triggered Safe Fallback.} When $\mathcal{R}$ exceeds threshold, Stage~2 overrides the command with a standing prompt and interpolates to a nominal pose. Without Stage~2 the robot collapses on the unstable reference; with it, the robot remains stable and awaits the next prompt.}
    \label{fig:safety_gating}
\end{figure}

\vspace{1.7mm}
\noindent\textbf{Stage~2\,\&\,3: Model- and Output-Level Filtering.}
Stage~1 filters semantically OOD prompts, but cannot detect internal generative instabilities, even under ID prompts. Thus, Stage~2 monitors the \emph{generation instability score} $\mathcal{R}$ online and triggers a fallback when the current reference becomes failure-prone.

\vspace{1.7mm}
\noindent\textit{(1) Is $\mathcal{R}$ a Meaningful Indicator?}
We validate $\mathcal{R}$ against downstream tracking error. We divide generated sequences into 10-frame windows, recording $\mathcal{R}$ and MPJPE ($\mathrm{mm}$), then group them into \emph{shared absolute} $\mathcal{R}$ quintiles across ID ({\small BABEL val., 220{,}944 windows}) and OOD ({\small Type B, 7{,}161 windows}) sequences, excluding windows after tracking failure. Figure~\ref{fig:instability_quintile} shows MPJPE increases monotonically with $\mathcal{R}$ in both domains, confirming that high-$\mathcal{R}$ references are harder to track. OOD windows also skew heavily toward the high-instability regime (Q5:\,$n{=}4{,}466$ vs. Q1:\,$n{=}178$). Importantly, this trend holds \emph{within ID alone}: even after Stage~1 acceptance, some ID windows fall into the high-instability regime (ID Q5: 87.4$\mathrm{mm}$), exhibiting larger errors than low-$\mathcal{R}$ OOD windows (OOD Q1: 56.6$\mathrm{mm}$). This shows Stage~1 alone is insufficient and motivates Stage~2.

We further validate the design of $\mathcal{R}$: as a fall predictor it outperforms the mean directional sensitivity $\bar{g}$ on matched rollouts (AUC 0.617 vs.\ 0.592), and $\mathcal{R}$ reaches an AUC of up to 0.899 with large-scale trackers. The F1-optimal threshold on evaluated rollouts is $\tau^*\!=\!4.9$---closely matching our deployed $\tau_{\mathrm{stab}}\!=\!5.0$---with a cross-tracker spread of only $\pm$0.3, indicating that the gate's operating point is stable across controllers.

\vspace{1.7mm}
\noindent\textit{(2) How Does $\mathcal{R}$ Behave During Streaming Generation?}
Figure~\ref{fig:safety_gating} (top, mid) shows $\mathcal{R}$ during streaming. For stable ID prompts (\eg, ``walk forward'', ``wave hands''), $\mathcal{R}$ remains low and smooth. Conversely, for extreme dynamics (``{\small Taekwondo}''), $\mathcal{R}$ sharply spikes, indicating unreliable flow dynamics and often preceding motion collapse.

\vspace{1.7mm}
\noindent\textit{(3) Instability Score-Triggered Safe Fallback:}
When $\mathcal{R}$ exceeds a threshold $\tau_{\mathrm{stab}}$, Stage~2 temporarily overrides the input with a safe ``stand'' prompt, interpolates the reference toward a predefined standing pose, and awaits the next input. Figure~\ref{fig:safety_gating} (bottom) shows the impact: without Stage~2 (w/o SFB), the tracker collapses; with Stage~2 enabled (w/ SFB), the system transitions to standing and remains stable.

Finally, Stage~3 performs deterministic output-level checks on joint position, velocity, and acceleration limits as the ultimate fail-safe before execution.

\begin{figure*}[t]
    \centering
    \includegraphics[width=\linewidth]{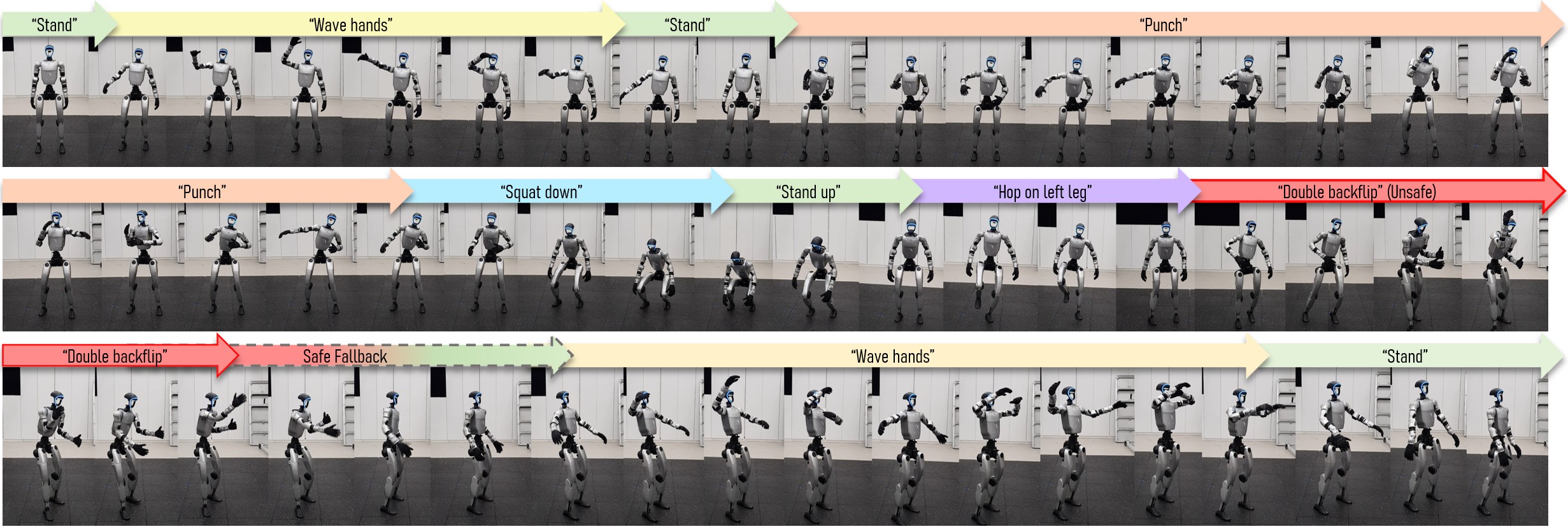}
    \vspace{-6mm}
    \caption{\textbf{Real-Robot Deployment on Unitree G1.} A continuous long-horizon sequence spanning upper-body gestures and whole-body actions; a high-risk prompt (``double backflip'') is filtered by the Safety Gate, triggering a standing fallback and allowing execution to continue---demonstrating sim-to-real transfer and deployment-time safety filtering.}
    \label{fig:real_robot}
    \vspace{-2mm}
\end{figure*}

\subsection{Real-Time Performance (Q3)}
\label{sec:exp_latency}
Since \textbf{SafeFlow} shares the text encoder~\cite{radford2021clip} and tracking controller with TextOp~\cite{xie2026textop}, we report generator and safety-gate latency only (RTX A6000, average of 100 runs). As Table~\ref{tab:latency} shows, physics-guided sampling increases latency, but \textit{Reflow} distillation reduces generation to $10.80\,\mathrm{ms}$. The 3-Stage Safety Gate adds $3.98\,\mathrm{ms}$ cumulatively, yielding $14.78\,\mathrm{ms}$ ($\sim\!67.7\,\mathrm{Hz}$) for the guarded generator. The shared text encoder ($\sim\!2.99\,\mathrm{ms}$) and ONNX-compiled controller ($\sim\!0.98\,\mathrm{ms}$) add little overhead, supporting real-time closed-loop control with asynchronous reference generation~\cite{xie2026textop}.

\begin{table}[t]
    \centering
    \caption{\textbf{Inference Latency and Pipeline Breakdown.} \textbf{SafeFlow} achieves real-time inference via reflow acceleration, while deployment-time safety gates add minimal overhead.}
    \vspace{-2mm}
    \label{tab:latency}
    \resizebox{\columnwidth}{!}{%
    \begin{tabular}{@{}lccc@{}}
        \toprule
        \textbf{Pipeline Component} & \textbf{Added\,(ms)$ \downarrow$} & \textbf{Latency\,(ms)$ \downarrow$} & \textbf{Freq.\,(Hz)$ \uparrow$} \\
        \midrule
        TextOp Generator~\cite{xie2026textop}               & - & 23.59\textsubscript{$\pm$1.60}    & 42.4 \\
        \midrule
        \textbf{SafeFlow} Generator (+\,Guid.)              & - & 172.03\textsubscript{$\pm$6.33}   & 5.8 \\
        \textbf{SafeFlow} Generator (+\,Guid.\,\&\,Reflow)    & - & 10.80\textsubscript{$\pm$0.98}    & 92.6 \\
        \quad + Stage 1 (Semantic OOD)                      & 0.006\textsubscript{$\pm$0.006}       & 10.81\textsubscript{$\pm$0.98}            & 92.5 \\
        \quad + Stage 2 (Generation Instability)            & 3.96\textsubscript{$\pm$1.04}         & 14.77\textsubscript{$\pm$1.43}            & 67.7 \\
        \quad + Stage 3 (Hard Kinematic Screen)             & 0.013\textsubscript{$\pm$0.003}       & \textbf{14.78}\textsubscript{$\pm$1.43}   & \textbf{67.7} \\
        \bottomrule
        \vspace{-2mm}
    \end{tabular}
    }
\end{table}

\subsection{Real-Robot Deployment (Q4)}
\label{sec:exp_real_robot}
We deploy \textbf{SafeFlow} on the Unitree G1 to evaluate sim-to-real transfer and on-hardware safety. As shown in Fig.~\ref{fig:real_robot}, the robot executes a continuous, long-horizon sequence---from upper-body gestures (``wave hands'', ``punch'') to demanding whole-body dynamics (``squat down'', ``hop on left leg'')---without intermediate stops or manual resets. Crucially, when a high-risk prompt (``double backflip'') known to induce unstable generation is issued mid-sequence, the safety gate filters the unsafe reference before it reaches the tracking controller and triggers the standing fallback, after which execution continues under subsequent prompts. These results demonstrate expressive long-horizon behaviors with enforced deployment-time safety on real hardware.

\begin{figure}[t]
    \centering
    \includegraphics[width=\columnwidth]{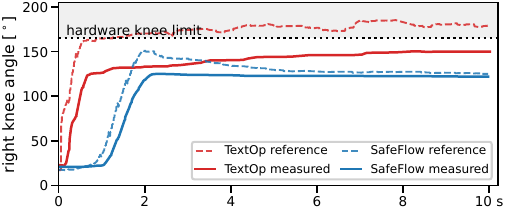}
    \vspace{-6mm}
    \caption{\textbf{Joint-Limit Infeasibility on Real Hardware} (``Squat down'', right knee). TextOp's reference commands the knee 10--20$^\circ$ beyond the hardware limit for the entire hold phase, which the robot cannot track; \textbf{SafeFlow}'s reference stays within limits and is tracked closely.}
    \label{fig:limit_causality}
    \vspace{-2mm}
\end{figure}

\smallskip\noindent\textbf{Quantitative Hardware Comparison.}
We further quantify hardware tracking fidelity by replaying each method's reference on the G1 for three commands (five 10\,s trials per cell) and measuring $E_{\mathrm{dof}}$ from the joint encoders against each method's \emph{own} reference (23 non-wrist joints, as in Table~\ref{tab:executability}). All executed trials completed without falls or operator intervention, and \textbf{SafeFlow} tracks more accurately on every command (16--52\% lower $E_{\mathrm{dof}}$; Table~\ref{tab:realrobot}). Since $E_{\mathrm{dof}}$ is measured against each method's own reference, low error cannot be obtained merely by conservative generation: on ``Squat down'', where the two references have comparable amplitude and mean joint speed (within 19\%), the gap persists. Moreover, the gap concentrates where TextOp's references leave the feasible range: on ``Squat down'', 18.0\% of its commanded joint samples lie outside the G1's hardware limits (Fig.~\ref{fig:limit_causality}), tracking error on those samples is 3.2$\times$ that of in-limit samples (21.4$^\circ$ vs.\ 6.7$^\circ$), and the three violating joints account for 88\% of the per-joint gap---the deployment-side counterpart of the JV statistics in Table~\ref{tab:executability}.

\begin{table}[t]
    \centering
    \caption{\textbf{Real-Robot Joint Tracking on Unitree G1.} $E_{\mathrm{dof}}$ (deg): encoder-measured MAE over the 23 non-wrist joints against each method's own reference (mean$\pm$std over 5 trials).}
    \vspace{-2mm}
    \label{tab:realrobot}
    \resizebox{0.75\columnwidth}{!}{
    \begin{tabular}{lcc}
        \toprule
        Command & TextOp~\cite{xie2026textop} & \textbf{SafeFlow (Ours)} \\
        \midrule
        \footnotesize``\textit{Wave hands}'' & 9.0\textsubscript{$\pm$0.2}  & \textbf{4.4}\textsubscript{$\pm$0.1} \\
        ``\textit{Squat down}'' & 9.3\textsubscript{$\pm$0.4}  & \textbf{6.1}\textsubscript{$\pm$0.2} \\
        ``\textit{Boxing}''     & 17.5\textsubscript{$\pm$0.8} & \textbf{14.6}\textsubscript{$\pm$0.2} \\
        \bottomrule
    \end{tabular}
    }
    \vspace{3mm}
\end{table}

\section{Conclusion}
\textbf{SafeFlow} advances real-time text-driven humanoid control toward safer deployment by addressing physically infeasible generations and OOD prompts through physics-guided generation and selective execution. Physics-Guided Rectified Flow with Reflow improves real-robot executability and real-time efficiency, while a 3-Stage Safety Gate filters semantic OOD inputs, generative instabilities, and kinematic violations before execution. Experiments on the Unitree G1 demonstrate improved physical feasibility, tracking success, and inference speed over diffusion- and retargeting-based baselines, with consistent gains across multiple downstream trackers. Overall, \textbf{SafeFlow} enables robust text-conditioned humanoid control under open-ended commands.

\balance
{
\footnotesize
\bibliographystyle{IEEEtran}
\bibliography{main}
}

\end{document}